\documentclass[runningheads]{llncs}
\usepackage{graphicx}
\usepackage{multirow}
\usepackage{algorithm}
\usepackage{algpseudocode}
\usepackage{makecell}
\usepackage{color, colortbl}
\usepackage{amsmath}
\usepackage{hyperref}
\definecolor{Gray}{gray}{0.9}
\definecolor{LightCyan}{rgb}{0.88,1,1}
\definecolor{amber}{rgb}{1.0, 0.75, 0.0}
\definecolor{sunset}{rgb}{0.98, 0.92, 0.84}

\begin{document}
%
\title{SEDA: Self-Ensembling ViT with Defensive Distillation and Adversarial Training for robust Chest X-rays Classification}
\titlerunning{SEDA}
%
\author{Raza Imam\and
Ibrahim Almakky\and
Salma Alrashdi\and
Baketah Alrashdi\and
Mohammad Yaqub}
%
\authorrunning{R. Imam et al.}
%
\institute{Mohamed Bin Zayed University of Artificial Intelligence, Abu Dhabi, UAE\\
\email{\{raza.imam, ibrahim.almakky, salma.alrashdi, baketah.alrashdi, mohammad.yaqub\}@mbzuai.ac.ae}}
%
\maketitle              
\begin{abstract}
Deep Learning methods have recently seen increased adoption in medical imaging applications. However, elevated vulnerabilities have been explored in recent Deep Learning solutions, which can hinder future adoption.
Particularly, the vulnerability of Vision Transformer (ViT) to adversarial, privacy, and confidentiality attacks raise serious concerns about their reliability in medical settings. This work aims to enhance the robustness of self-ensembling ViTs for the tuberculosis chest x-ray classification task. We propose \underline{S}elf-\underline{E}nsembling ViT with defensive \underline{D}istillation and \underline{A}dversarial training (SEDA).  SEDA utilizes efficient CNN blocks to learn spatial features with various levels of abstraction from feature representations extracted from intermediate ViT blocks, that are largely unaffected by adversarial perturbations. Furthermore, SEDA leverages adversarial training in combination with defensive distillation for improved robustness against adversaries.
Training using adversarial examples leads to better model generalizability and improves its ability to handle perturbations. Distillation using soft probabilities introduces uncertainty and variation into the output probabilities, making it more difficult for adversarial and privacy attacks. 
Extensive experiments performed with the proposed architecture and training paradigm on publicly available Tuberculosis x-ray dataset shows SOTA efficacy of SEDA compared to SEViT in terms of computational efficiency with $70\times$ times lighter framework and enhanced robustness of +9\%. Code: \href{https://github.com/Razaimam45/SEDA}{Github}.

\keywords{Ensembling \and Adversarial Attack \and Defensive Distillation \and Adversarial Training \and Vision Transformer}
\end{abstract}

\section{Introduction}

Deep Learning (DL) models have proven their efficacy on various medical tasks in general and on classifying lung diseases from x-ray images in particular \cite{ccalli2021deep}. 
However, popular DL models, such as Convolutional Neural Networks (CNNs) and Vision Transformers (ViTs) are susceptible to adversarial attacks. The vulnerabilities of medical imaging systems to such attacks are increasingly apparent \cite{goodfellow2014explaining}. These attacks can target cloud-based processing or communication channels to compromise patient data \cite{huang2022privacy}. Adversarial attacks in medical imaging can also be exploited for financial gain through fraudulent billing or insurance claims \cite{almalik2022self}. Moreover, malicious parties extracting model information could compromise patient data privacy. Therefore, it is essential to develop robust defense strategies to ensure the secure deployment of automated medical imaging systems \cite{imam2023enhancing}. Detecting and defending against adversarial attacks is crucial for accurate diagnoses and successful treatment outcomes in the healthcare domain. \cite{kaviani2022adversarial}.


Recently, SEViT \cite{almalik2022self}, a self-ensembling approach based on Vision Transformers, has shown promise in defending against adversarial attacks in medical image analysis. However, SEViT has practical deployment limitations due to the large parameter size of the MLP modules ensembled for each block.
Attacks such as model extraction can extract SEViT models, despite their effectiveness against adversarial attacks \cite{carlini2023extracting}. Privacy attacks, such as model extraction in the medical imaging domain, pose significant threats to patient privacy and confidentiality. For instance, attackers can extract trained models from healthcare facilities and exploit them for malicious purposes, including selling the models, training their own models using the extracted models, or inferring sensitive patient information from x-ray images.


Medical data can contain sensitive information about patients, and unauthorized model extraction can lead to privacy violations \cite{rasool2022security}. 
To defend against such model extraction attacks, defensive distillation \cite{papernot2016distillation} can be employed. 
Defensive distillation enhances model robustness by training a distilled model on softened probabilities from an initial model, mitigating adversarial attacks by introducing prediction uncertainty, and reducing sensitivity to small perturbations \cite{imam2023enhancing}. Soft probabilities refer to the output of a model's prediction that represents the likelihood or probability distribution over multiple classes rather than a single deterministic label. 
By training a distilled model that approximates the behavior of the original model, we can enhance resistance to extraction. By making extraction more challenging, attackers may be deterred due to the increased time and resources required.
In this work, we aim to improve the robustness of the SEViT model for Tuberculosis (TB)  classification from chest x-ray images using defensive distillation and adversarial training, while also improving computational efficiency. The main aim of this research is to provide a lightweight, accurate, and robust benchmark medical imaging system, which aims to be crucial in achieving better diagnosis and treatment outcomes for patients and healthcare providers.
The main contributions of this work are as follows:
\begin{itemize}
    \item We propose an efficient and lightweight \underline{S}elf-\underline{E}nsembling ViT with Defensive \underline{D}istillation and \underline{A}dversarial training (SEDA)
    to increase adversarial robustness while preserving the clean accuracy of ViT by combining the adversarial training and defensive distillation approaches.
    \item Showing the real-world deployability of SEDA, we evaluate the proposed framework's computational parameters and classification performance, and compare them with state-of-the-art methods \cite{vaswani2017attention} and \cite{almalik2022self}. 
\end{itemize}


\section{Related Works}
\textbf{Ensembling ViTs.} 
Adversarial attacks can be more potent against ViTs by targeting both intermediate representations and the final class token \cite{naseer2021improving}. To address this challenge, \cite{almalik2022self} proposed the Self-Ensemble Vision Transformer (SEViT) architecture. SEViT enhances the robustness of ViT against adversarial attacks in medical image classification. They propose to add an MLP classifier at the end of each block to leverage patch tokens and generate a probability distribution over class labels. This approach enables the self-ensembling of classifiers, which can be fused to obtain the final classification result.

\textbf{Adversarial Attacks.} The works by \cite{Malik_2022_BMVC,goodfellow2014explaining,wu2022towards} analyzed a common theme of exploring adversarial attacks in machine learning. They aim to better understand the nature of these attacks and propose new techniques for mitigating their effects on learning models. \cite{Malik_2022_BMVC} presents a new technique for creating adversarial examples, while \cite{goodfellow2014explaining} proposes a defense mechanism against them. \cite{wu2022towards} presents an efficient approach to training models to resist adversarial attacks.

\textbf{Adversarial Defense.} \cite{papernot2016distillation} proposes a defense mechanism called defensive distillation, which is a form of knowledge transfer from a larger, more accurate model to a smaller, less accurate model. The idea of \cite{papernot2016distillation} is to make the smaller model less vulnerable to adversarial attacks by training it to imitate the outputs of the larger model, which is assumed to be more robust to adversarial perturbations. 

\section{Proposed Method}

\begin{figure}[t]
    \centering\includegraphics[width=0.8\textwidth]{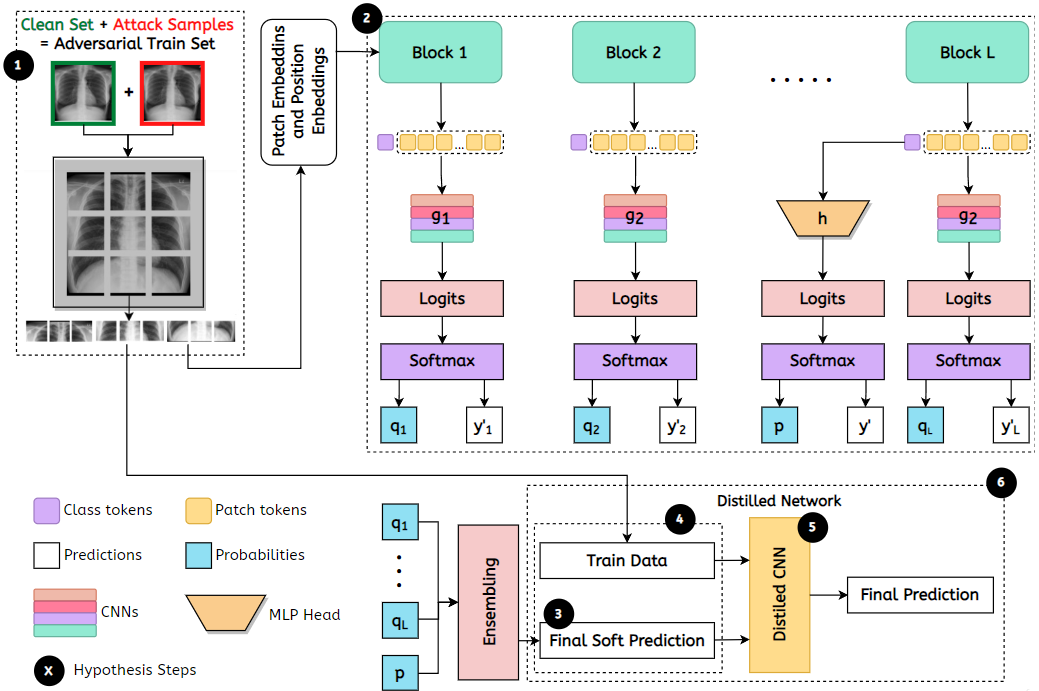}
    \caption{The proposed SEDA framework extracts the patch tokens from the initial
    blocks and trains separate CNNs on Clean+Attack samples (Adversarial Training) as shown in (1) and (2). A self-ensemble of these CNNs with the final ViT classifier goes through the distillation process with the new adversarial trained dataset to obtain a final distilled CNN model as seen in steps (3), (4), (5), and (6). }
    \label{fig:method}
\end{figure}

The objective of an adversarial attack is to generate a perturbed image $x'$, which is similar to the original medical image $x$ within a certain distance metric ($L_{\infty}$ norm), such that the output of a ViT-based classifier $f(x')$ is different from the true label $y$ with a high probability. Defending against adversarial attacks involves obtaining a robust classifier $f'$ from the original classifier $f$. This robust classifier $f'$ should have high accuracy on both clean images ($P(f'(x) = y)$) and perturbed images ($P(f'(x') = y)$). 

\textbf{Ensembling} SEViT \cite{almalik2022self} aims to improve adversarial robustness by adding an MLP classifier at the end of first $m$ blocks, utilizing patch tokens to produce a probability distribution over class labels. The intermediate feature representations output by the initial blocks are considered useful for classification and more robust against adversarial attacks.
This results in a self-ensemble of $L$ classifiers that can be fused to obtain the final classification result. The SEViT model 
increases adversarial robustness by adding MLP classifiers only to the first $m$ ($m < L$) blocks and combining their results with the final classification head. The SEViT ensemble can be formed by performing majority voting or randomly choosing only $c$ out of the initial $m$ intermediate classifiers. The SEViT defense mechanism also includes a detection mechanism, which aims to distinguish between the original image $x$ and the perturbed image $x'$, especially when the attack is successful ($f(x') \neq y$). The constraint on the distance metric between $x$ and $x'$ is that the $L_{\infty}$ norm of their difference should be less than or equal to a predefined value epsilon. Although, we are not proposing any detection mechanism (to detect adversarial samples) in our enhanced solution as we can use the same detection approach as SEViT.

\textbf{SEDA.} Our two main hypotheses are: (1) Small CNN blocks would be more computationally efficient alternative to MLP blocks for each ViT block. (2) Defensive distillation when performed with adversarial training would make the model more robust against adversarial and model extraction attacks. Hence, we propose to enhance the existing SEViT by making 3 major modifications: We propose to substitute MLP blocks with CNN blocks instead to test for efficiency. Next, we perform adversarial training on the SEViT model instead of training it on just clean samples. Finally, we generate soft predictions to train a new distilled model, resulting in the SEDA framework (depicted in Figure \ref{fig:method}).

The proposed modifications are based on the fact that CNNs are more efficient than MLPs in learning spatial features from tokens through convolution operations at different levels of abstraction, which leads to improved generalization performance and reduced overfitting. Additionally, by training the model with adversarial examples, the model becomes better at handling perturbations during inference and generalizes better to new and unseen adversarial examples. The use of soft probabilities during distillation results in a smaller and more efficient model with faster inference time and reduced computational requirements. Moreover, the distilled model is more robust to adversarial attacks since the soft probabilities introduce uncertainty and variation in the output probabilities, making it harder for attackers to generate adversarial examples that can fool the model. 

\section{Experiments}

\textbf{Dataset.} The experiments are performed on a chest x-ray dataset \cite{rahman2020reliable} that includes $7,000$ images, and the classification task is binary, where each image is either Normal or TB. The dataset was split randomly, with $80\%$ allocated for training, $10\%$ for validation, and the remaining $10\%$ for testing. This split ensures that the training, validation, and testing sets are mutually exclusive, and it enables the evaluation of the model's generalization ability to unseen images.

\textbf{Attack Types.} In our study, we utilize the Foolbox library \cite{rauber2020foolbox} to create $3$ different types of adversarial attacks,  which are FGSM \cite{goodfellow2014explaining}, PGD \cite{madry2017towards}, and AutoPGD \cite{croce2020reliable}. To generate attack samples with these algorithms, we use two values of perturbation budget ${\epsilon}=0.03$ and ${\epsilon}=0.003$ while keeping all other parameters at their default values.
  
\textbf{MLPs alternatives.} To implement the Vision Transformer (ViT) model, we utilized the ViT-B/16 architecture pre-trained on ImageNet [5]. We used the fine-tuned ViT used in the original SEViT method. We experimented with several MLP alternatives with the aim to preserve clean accuracy and enhance robust accuracy while significantly reducing computational requirements. In order to create intermediate classifiers that take patch tokens as input, we trained 12 MLP alternatives such as CNN and ResNet variants for every block.  
In particular, we trained alternatives including a 2 convolution layer CNN, fine-tuned ResNet-50, transfer-learned ResNet-50, fine-tuned ResNet-50 with 2 additional convolution layers, and transfer-learned ResNet-50 with 2 additional convolution layers. Fine-tuning involves training the ResNet-50 model on our specific dataset, while transfer learning involves using the pre-trained weights from ImageNet and adapting the model to our specific task. 
Our experiments were conducted on a single Nvidia Quadro RTX 6000 GPU.  All models are tested with the batch size of $30$ and input size of $3 \times 224 \times 224$.

\begin{figure}[b]
\centering
\begin{minipage}{.50\textwidth}
  \centering
  \includegraphics[width=6cm]{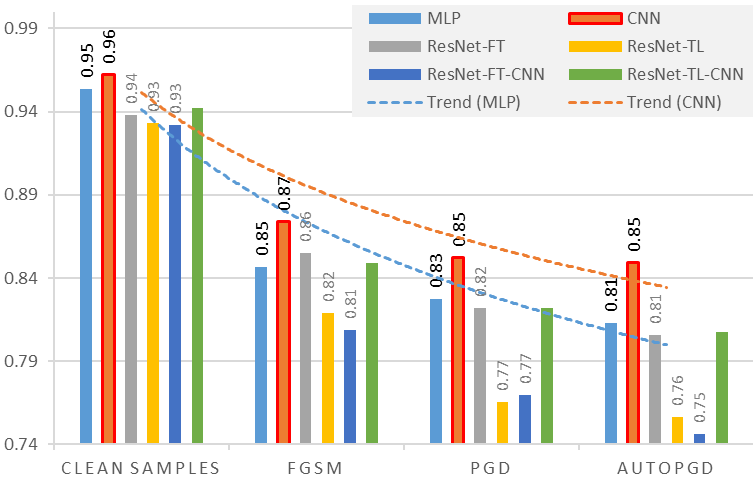}
\end{minipage}%
\begin{minipage}{.50\textwidth}
  \centering
  \includegraphics[width=6cm]{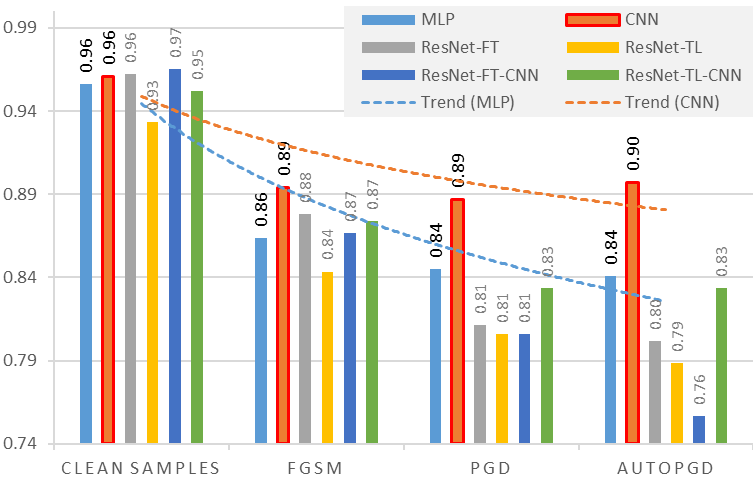}
\end{minipage}
\caption{Clean and adversarial performance of MLP alternatives Before adversarial training (left) vs After adversarial training (right). For each MLP alternative, the number of ensembles are m=3 where the attack samples have ${\epsilon=0.03}$}
\label{fig:adv_performance}

\end{figure}

\section{Results and Discussion}

\textbf{Accuracy and Robustness.} 
Comparing the performances (Table \ref{tab:vit_vs_mlp_vs_cnn} and Figure \ref{fig:adv_performance}) of our approach to the  MLP block while ensembling consistently on 3 models (i.e., m=3), as illustrated in Figure \ref{fig:adv_performance}, we conclude that the 2 convolution layer CNN alternative outperforms the original MLP block as it closely preserves the clean accuracy (with mean difference of $\le1\%$) and achieves higher accuracy against adversarial attack samples and that with significantly lower number of parameters. This can be attributed to CNN's ability to learn spatial features from tokens through convolution operations at different levels of abstraction, leading to reduced overfitting and improved generalization performance. For example, when subjected to attack samples generated by FGSM ($\epsilon$=0.03), the CNN alternative achieved $87.4\%$ robust accuracy, which is higher than the $84.6\%$ robust accuracy achieved by the 4-layered MLP block. When combined with distillation and adversarial training, the proposed SEDA framework is seen to achieve even higher clean and robust accuracy, exceeding an improvement of $6\%$. This increment in robust accuracy can also be noticed consistently on other attacks.

\textbf{Computation.} In contrast to SEViT, our proposed SEViT-CNN, and its distilled version, SEDA, offer significantly improved clean and robust accuracies for real-world deployment. SEDA is also 70 times more memory efficient than the original SEViT, as demonstrated by the computational and accuracy parameters shown in Table \ref{tab:computation}.
The computational efficiency of a DL model can be evaluated based on various factors such as the FLOPs and accuracy. While the original MLP block in SEViT is accurate, it requires a significant amount of computational resources to train and deploy with having about $625$M parameters \cite{zhou2021decentralized} (Table \ref{tab:computation}).
Based on these findings, we conclude that the CNN alternative is the best choice for ensembling ViT considering clean and robust accuracy, as well as computational efficiency.

{\renewcommand{\arraystretch}{1.2}
\begin{table}[t]
\centering
\caption{ViT vs SEViT vs SEViT-CNN (highlighted) in terms of Clean and Robust accuracy across the different numbers of intermediate ensembles. m = \#Ensembles}
\label{tab:vit_vs_mlp_vs_cnn}
\resizebox{\textwidth}{!}{%
\begin{tabular}{lllllllll@{\hspace{3mm}}lllllll}
 &
   &
  \multicolumn{7}{l}{\textit{(a). PRE-Adversarial   Training}} &
  \multicolumn{7}{l}{\textit{(b). POST-Adversarial   Training}} \\ \hline
\multirow{2}{*}{Ensemble} &
  \multirow{2}{*}{m} &
  \multirow{1}{*}{Clean} &
  \multicolumn{2}{c}{FGSM} &
  \multicolumn{2}{c}{PGD} &
  \multicolumn{2}{c}{AutoPGD} &
  \multirow{1}{*}{Clean} &
  \multicolumn{2}{c}{FGSM} &
  \multicolumn{2}{c}{PGD} &
  \multicolumn{2}{c}{AutoPGD} \\ 
  \cline{4-9} \cline{11-16}
 &
   &
   &
  0.03 &
  0.003 &
  0.03 &
  0.003 &
  0.03 &
  0.003 &
   &
  0.03 &
  0.003 &
  0.03 &
  0.003 &
  0.03 &
  0.003 \\ \hline
{ViT \cite{vaswani2017attention}} &
  - &
  96.38 &
  55.65 &
  91.59 &
  32.62 &
  92.17 &
  23.77 &
  92.46 &
  - &
  - &
  - &
  - &
  - &
  - &
  - \\ \hline
\multirow{4}{*}{MLP \cite{almalik2022self}} &
  1 &
  94.20 &
  70.29 &
  90.44 &
  62.90 &
  91.16 &
  58.41 &
  91.59 &
  91.45 &
  70.15 &
  88.41 &
  62.61 &
  88.99 &
  58.70 &
  89.71 \\
 &
  2 &
  \textbf{96.52} &
  83.91 &
  95.22 &
  80.15 &
  95.36 &
  78.41 &
  \textbf{95.51} &
  95.51 &
  84.06 &
  93.48 &
  81.01 &
  93.62 &
  80.87 &
  93.91 \\
 &
  3 &
  95.36 &
  84.64 &
  94.06 &
  82.75 &
  93.91 &
  81.30 &
  94.20 &
  95.65 &
  86.38 &
  93.62 &
  84.49 &
  93.77 &
  84.06 &
  94.20 \\ \hline
\multirow{4}{*}{\makecell{CNN\\(Ours)}} &
  1 &
  93.04 &
  70.87 &
  89.86 &
  64.78 &
  90.58 &
  60.44 &
  90.73 &
  95.51 &
  71.30 &
  92.03 &
  63.77 &
  92.75 &
  58.84 &
  92.75 \\
 &
  2 &
  96.23 &
  85.22 &
  94.20 &
  82.32 &
  94.49 &
  82.03 &
  94.64 &
  95.65 &
  87.83 &
  94.64 &
  86.09 &
  94.78 &
  87.54 &
  94.64 \\
 &
  \cellcolor{LightCyan}3 &
  \cellcolor{LightCyan}96.23 &
  \cellcolor{LightCyan}\textbf{87.39} &
  \cellcolor{LightCyan}\textbf{94.64} &
  \cellcolor{LightCyan}\textbf{85.22} &
  \cellcolor{LightCyan}\textbf{94.93} &
  \cellcolor{LightCyan}\textbf{84.93} &
  \cellcolor{LightCyan}94.93 &
  \cellcolor{LightCyan}\textbf{96.09} &
  \cellcolor{LightCyan}\textbf{89.42} &
  \cellcolor{LightCyan}\textbf{95.36} &
  \cellcolor{LightCyan}\textbf{88.70} &
  \cellcolor{LightCyan}\textbf{95.36} &
  \cellcolor{LightCyan}\textbf{89.71} &
  \cellcolor{LightCyan}\textbf{95.22} \\ 
  \hline
\end{tabular}%
}
\end{table}}

{\renewcommand{\arraystretch}{1.2}
\begin{table}[t]
\centering
\caption{Comparative performance of different ensembling models and their distilled versions (with \#ensembles (m) = 3) against clean samples and attack samples. The reported results are the median of multiple runs.
}
\label{table:pre_dis}
\resizebox{\textwidth}{!}{%
\begin{tabular}{llllllll@{\hspace{3mm}}lllllll}
\multirow{3}{*}{Model} &
  \multicolumn{7}{l}{\textit{(a). PRE-Adversarial   Training}} &
  \multicolumn{7}{l}{\textit{(b). POST-Adversarial   Training}} \\ \hline
 &
  \multirow{2}{*}{Clean} &
  \multicolumn{2}{c}{FGSM} &
  \multicolumn{2}{c}{PGD} &
  \multicolumn{2}{c}{AutoPGD} &
  Clean &
  \multicolumn{2}{c}{FGSM} &
  \multicolumn{2}{c}{PGD} &
  \multicolumn{2}{c}{AutoPGD} \\ \cline{3-8} \cline{10-15}
                      &       & 0.03  & 0.003 & 0.03  & 0.003 & 0.03  & 0.003 &       & 0.03  & 0.003 & 0.03  & 0.003 & 0.03  & 0.003 \\ \hline
ViT (m=0)             & \textbf{96.38} & 55.65 & 91.59 & 32.62 & 92.17 & 23.77 & 92.46 & -     & -     & -     & -     & -     & -     & -     \\
SEViT           & 95.36 & 84.64 & 94.06 & 82.75 & 93.91 & 81.30 & 94.20 & 95.65 & 86.38 & 93.62 & 84.49 & 93.77 & 84.06 & 94.20 \\
SEViT-CNN (Ours)       & 96.23 & 87.39 & \textbf{94.64} & 85.22 & 94.93 & 84.93 & 94.93 & \textbf{96.09} & 89.42 & 95.36 & 88.70 & 95.36 & 89.71 & 95.22 \\
Distilled (ViT)       & 94.71 & 84.64 & 87.10 & 86.67 & 93.33 & \textbf{90.87} & 94.20 & -     & -     & -     & -     & -     & -     & -     \\
Distilled (SEViT)     & 92.71 & 84.78 & 88.26 & 86.52 & 93.33 & 88.70 & 93.91 & 93.86 & 87.39 & \textbf{97.68} & 87.83 & \textbf{97.97} & 90.87 & \textbf{96.67} \\

SEDA (Ours) & 94.43 & \textbf{88.84} & 94.20 & \textbf{88.70} & \textbf{95.65} & 90.29 & \textbf{95.94} & \cellcolor{LightCyan} 94.86 & \cellcolor{LightCyan}\textbf{89.86} & \cellcolor{LightCyan}96.38 & \cellcolor{LightCyan}\textbf{90.87} & \cellcolor{LightCyan}96.09 & \cellcolor{LightCyan}\textbf{92.46} & \cellcolor{LightCyan}96.09 \\ \hline
\end{tabular}%
}
\end{table}}

{\renewcommand{\arraystretch}{1.2}
\begin{table}[t]
\centering
\caption{Comparison of extracted model when model extraction is performed on original model V/s when extraction is performed on the distilled versions (with m=3). From defender’s view, \textit{Lower} clean/Adv. accuracy is better. Note, SEDA (highlighted) = SEViT-CNN + Distillation + Post Adversarial Training.}
\label{table:pos_ext}
\resizebox{\textwidth}{!}{%
\begin{tabular}{llllllll@{\hspace{3mm}}lllllll}
\multirow{3}{*}{Extraction On} &
  \multicolumn{7}{l}{\textit{(a). PRE-Adversarial   Training}} &
  \multicolumn{7}{l}{\textit{(b). POST-Adversarial   Training}} \\ \hline
 &
  \multirow{2}{*}{Clean} &
  \multicolumn{2}{c}{FGSM} &
  \multicolumn{2}{c}{PGD} &
  \multicolumn{2}{c}{AutoPGD} &
  Clean &
  \multicolumn{2}{c}{FGSM} &
  \multicolumn{2}{c}{PGD} &
  \multicolumn{2}{c}{AutoPGD} \\ \cline{3-8} \cline{10-15}
                      &       & 0.03  & 0.003 & 0.03  & 0.003 & 0.03  & 0.003 &       & 0.03  & 0.003 & 0.03  & 0.003 & 0.03  & 0.003 \\ \hline
ViT (m=0)             & 84.57 & 82.75 & 85.65 & 82.61 & 85.65 & 83.77 & 85.94 & -     & -     & -     & -     & -     & -     & -     \\
SEViT           & 84.43 & 83.62 & 85.51 & 83.04 & 85.51 & 83.62 & 85.51 & 83.57 & 82.32 & 84.93 & 82.90 & 84.93 & 82.46 & 85.07 \\
SEViT-CNN (Ours)       & 84.57 & 81.30 & 84.64 & 81.74 & 84.78 & 82.32 & 85.36 & \textbf{81.43} & \textbf{80.43} & \textbf{82.61} & \textbf{80.29} & \textbf{82.75} & \textbf{80.14} & \textbf{82.90} \\
Distilled (ViT)       & 85.14 & 83.04 & 85.65 & 82.32 & 85.80 & 82.32 & 86.23 & -     & -     & -     & -     & -     & -     & -     \\
Distilled (SEViT)     & 83.57 & 78.99 & \textbf{83.77} & \textbf{79.13} & 84.60 & 79.75 & 84.64 & 84.57 & 82.90 & 85.36 & 82.32 & 85.36 & 83.19 & 85.80 \\
SEDA (Ours) & \textbf{83.71} & \textbf{79.86} & 84.20 & 79.71 & \textbf{84.20} & \textbf{79.71} & \textbf{84.64} & \cellcolor{LightCyan} \textbf{81.86} & \cellcolor{LightCyan}\textbf{82.03} & \cellcolor{LightCyan}\textbf{83.04} & \cellcolor{LightCyan}\textbf{82.17} & \cellcolor{LightCyan}\textbf{83.04} & \cellcolor{LightCyan}\textbf{82.03} & \cellcolor{LightCyan}\textbf{83.48} \\ \hline
\end{tabular}%
}
\end{table}}

\subsection{Defensive Distillation vs Extraction Attack}
\label{sec:defense_vs_attack}

\textbf{Distillation.} 
Table \ref{table:pre_dis} indicates that the distilled model outperformed the original model in terms of robust accuracy despite having a smaller architecture (of 5 convolution layer CNN). However, there was a slight decline of around 2\% in the clean accuracy of the distilled model, which could be viewed as a reasonable trade-off between clean and robust accuracy. Distillation using soft probabilities leads to a compact and smaller architecture, which results in faster inference time and lower computational requirements, making the model more efficient as shown in Table \ref{tab:computation}. Furthermore, the distilled model is less susceptible to adversarial attacks as soft probabilities impart a smoothing effect during the distillation process \cite{carlini2017towards}. This is because soft probabilities introduce some degree of uncertainty and variation in the output probabilities of the model, making it more difficult for an attacker to generate adversarial examples that can fool the model. Hence, the distilled model, SEDA, is robust against adversarial attacks than the original SEViT-CNN with a slight trade-off with clean accuracy.

\textbf{Extraction.} We evaluate the performance of original and distilled models against extraction attacks in a black-box setting where the attacker can only input queries and receive outputs. The attacker utilizes this input-output relationship to create a replica/attack/extracted model (a 3 convolution layer CNN) of the original model. The aim is to compare the extracted model's performance in two scenarios: one where the original model is deployed, and the other where the distilled variant is deployed. This comparison allows us to determine that deploying which model would be more resilient to model extraction attacks.
Table \ref{table:pos_ext} shows that the attacker's clean accuracy (post-adversarial training) in reproducing the distilled SEViT-CNN, i.e., SEDA (81.86\%) is lower compared to that of the original SEViT-CNN (84.57\%) and SEViT (84.43\%). The models extracted by the attacker, exhibit lower level of accuracy and robustness than the original model, with SEViT-CNN and SEDA being among the lowest, implying that the attacker is less likely to recover the exact parameters of these models due to the smoothed probabilities. Thus, we can conclude that deploying the distilled model is a more secure option against extraction attacks than the original SEViT or ViT model.


\subsection{Adversarial Training}
\label{sec:adv_train}

\noindent \textbf{Accuracy and Robustness.} In the case of pre-adversarial training, both models (Original and Distilled) have high accuracy on clean samples, but their performance on adversarial examples is increased even higher following post-adversarial training, by at least +2 to +5\% (Table \ref{tab:vit_vs_mlp_vs_cnn} and Table \ref{table:pre_dis}). Additionally, the models' vulnerability to model extraction attacks is high in this pre-adversarial training case, with the extracted models' accuracy on clean and adversarial examples being considerably high. However, following post-adversarial training, models extracted on original and their distilled versions showed a considerable decrement (of about -3\%) in their clean and robust accuracies (Table \ref{table:pos_ext}). These results demonstrate the improved robustness of the models via adversarial training with improved generalization and their increased resistance against attacks and model extraction.

{\renewcommand{\arraystretch}{1.2}
\begin{table}[t]
\centering
\caption{Extensive computational analysis of the SEDA and its comparison with the alternative models in terms of computational parameters. \textit{Distilled} models are distilled with \#ensembles (m) = 3. The '+' refers to \textit{in addition} to ViT parameters. 
}
\label{tab:computation}
\resizebox{\textwidth}{!}{%
\begin{tabular}{lllllllllll}
\hline
\multirow{2}{*}{Model} &
  \multirow{2}{*}{\#Params} &
  \multirow{2}{*}{FLOPs} &
  \multirow{2}{*}{Weight} &
  \multirow{2}{*}{\makecell{\text{Inference}\\ \text{Time}}} &
  \multirow{2}{*}{Throughput} &
  \multirow{2}{*}{\makecell{\text{Clean}\\ \text{Accuracy}}} &
  \multicolumn{4}{c}{Adv. Acc. ($\epsilon=0.03$)} \\ \cline{8-11}
                      &          &          &          &       &           &       & FGSM  & PGD   & AutoPGD & Mean\\ \hline
ViT (m=0)             & 85.64M   & 16.86G   & 327.37MB & 7.06  & 277.25    & \textbf{96.38} & 55.65 & 32.62 & 23.77 & 37.35   \\
\cellcolor{sunset}SEViT (m=3)           & \cellcolor{sunset}+1875.66M & \cellcolor{sunset}+1875.63M  & \cellcolor{sunset}+6.99GB   & \cellcolor{sunset}12.87 & \cellcolor{sunset}13125.57  & \cellcolor{sunset}\textbf{95.36} & \cellcolor{sunset}84.64 & \cellcolor{sunset}82.75 & \cellcolor{sunset}81.30 & \cellcolor{sunset}82.90  \\
\cellcolor{LightCyan}SEViT-CNN (m=3)       & \cellcolor{LightCyan}+3.09M    & \cellcolor{LightCyan}+531.27M  & \cellcolor{LightCyan}+11.82MB  & \cellcolor{LightCyan}1.20  & \cellcolor{LightCyan}124761.51 & \cellcolor{LightCyan}\textbf{96.23} & \cellcolor{LightCyan}87.39 & \cellcolor{LightCyan}85.22 & \cellcolor{LightCyan}84.93 & \cellcolor{LightCyan}85.85  \\
Distilled (ViT)       & 27.79M   & 994.41M  & 106.00MB & 0.96  & 3956.04   & 94.71 & 84.64 & 86.67 & 90.87 & 87.39  \\
Distilled (SEViT)     & 27.79M   & 994.41M  & 106.00MB & 0.97  & 3952.44   & 92.71 & 84.78 & 86.52 & 88.70 & 86.67  \\
\cellcolor{LightCyan}SEDA (Ours) & \cellcolor{LightCyan}\textbf{27.79M}   & \cellcolor{LightCyan}\textbf{994.41M}  & \cellcolor{LightCyan}\textbf{106.00MB} & \cellcolor{LightCyan}0.96  & \cellcolor{LightCyan}3962.13   & \cellcolor{LightCyan}\textbf{94.86} & \cellcolor{LightCyan}\textbf{89.86} & \cellcolor{LightCyan}\textbf{90.87} & \cellcolor{LightCyan}\textbf{92.46} & \cellcolor{LightCyan}\textbf{91.06} \\ \hline
\end{tabular}
}
\end{table}}

\section{Conclusion and Future Work}
In this work, we have proposed an improved architecture, Self-Ensembling ViT with defensive Distillation and Adversarial training (SEDA), to overcome computational bottlenecks and improve robustness by making 3 major modifications: we propose to substitute MLP blocks with CNN instead to test for efficiency. Next, we perform adversarial training on the SEViT-CNN model instead of training it on just clean samples. And last, we generate soft predictions to train a new distilled model. This study is the first to combine these techniques with ViT ensembling, providing a SOTA defense for ViTs against adversarial attacks and model extraction attacks. We prove the effectiveness of our novel enhanced approach: SEDA, using an extensive set of experiments on publicly available Tuberculosis x-ray dataset. 

In the future, we aim to further enhance the proposed SEDA model using (i) defensive approaches such as differential privacy (ii) exploring the trade-off between adversarial robustness and model accuracy when performing defensive distillation (iii) exploring a diverse set of alternatives for each block rather than just having CNN for every block. This would increase the diversity among the ensemble models and further improve overall performance. 

\bibliographystyle{unsrt} 
\bibliography{references} 

\clearpage

\section*{Supplementary}

\begin{algorithm}[h]
\caption{Pseudocode of the proposed SEDA framework\\ \textbf{Input:} Batch of clean/attack TB images;
\textbf{Output:} Robust predictions;}
\label{alg:method}
\begin{algorithmic}[1]
\State Define the SEViT model architecture with CNN blocks instead of MLP blocks. Below, $\psi$ = Majority Voting ensembling, $\theta$ = parameters of the final ViT classifier $\mathbf{h}$, and $\beta_1$ through $\beta_m$ = parameters of the m intermediate CNNs g. \newline \vspace{-0.2cm}
\[\tilde{\mathbf{f}}(\mathbf{x})=\psi\left(\mathbf{g}_{\beta_1}\left(\left\{p t_1^j\right\}_{j=1}^N\right), \cdots, \mathbf{g}_{\beta_m}\left(\left\{p t_m^j\right\}_{j=1}^N\right), \mathbf{h}_\theta\left(c t_L\right)\right)\]
\vspace{-0.2cm}

\State Train the SEViT-CNN model on the original + adversarial datasets, i.e., adversarial training. Below, ${L}$ and ${L'}$ = original and adversarial loss on clean $x_i$ and adversarial samples $x_i'$ respectively. \newline \vspace{-0.5cm}
\[\text{Adversarial Training Loss} = \frac{1}{N} \sum_{i=1}^{N} \left[ \lambda \cdot {L}(\tilde{\mathbf{f}}(x_i, y_i)) + (1 - \lambda) \cdot {L'}(\tilde{\mathbf{f}}(x_i', y_i)) \right]\]\vspace{-0.2cm}

\State Extract the final soft predictions from the Adversarially trained SEViT-CNN model $\tilde{\mathbf{f'}}(\cdot)$ for all images in the dataset.\newline \vspace{-0.3cm}
\[
\text{Soft Predictions} = \{y_1', y_2', y_3', \ldots, y_m\} = \tilde{\mathbf{f'}}(x_1, x_2, x_3, \ldots, x_m)
\]

\State Create a new dataset consisting of the images from both the original and adversarial datasets along with their corresponding soft predictions. Below, $y_i'$ is soft prediction. \vspace{-0.3cm}
\[\text{Data Point} = [X:\{x_1, x_2, x_3, \ldots, x_m\}, Y:\{y_1', y_2', y_3', \ldots, y_m\}]\] 

\State Train a new distilled model (such as CNN) on the new dataset, where each data point consists of an image and its corresponding soft predictions. Below, $z_i$ = the activation before the softmax, $T$ = temparature parameter. \vspace{-0.2cm}

\[\mathbf{F(x)}=\left[\frac{e^{z_i(X) / T}}{\sum_{l=0}^{N-1} e^{z_l(X) / T}}\right]_{i \in 0 . . N-1}\]

\end{algorithmic}
\end{algorithm}

\begin{figure}[htpb]
    \centering
    \includegraphics[width=6cm]{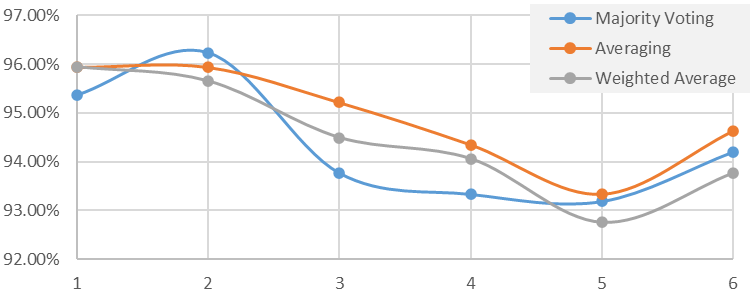}
    \caption{Comparison of clean performance of CNN blocks when ensembling is done with different voting criterias in SEDA. The x-axis is the number of ensembles while y-axis is the test accuracy}
    \label{fig:voting}
\end{figure}

\begin{figure}[htpb]
    \centering
    \includegraphics[width=6cm]{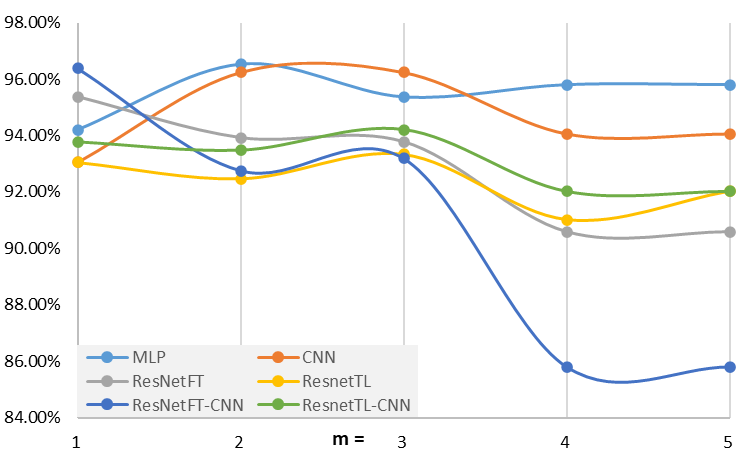}
    \caption{Comparison of clean performance of different MLP alternatives across various ensembles}
    \label{fig:CNN&MLP}
\end{figure}

{\renewcommand{\arraystretch}{1.2}
\begin{table}[htpb]
\centering
\caption{Clean performance of different MLP alternatives across several ensembles using Majority Voting. The `+' refers to \textit{in addition} to ViT parameters.}
\label{table:clean_with_params}
\resizebox{8cm}{!}{%
\begin{tabular}{llllllll}
\hline
Ensemble &
  \#Params & m=1 & m=2 & m=3 & m=4 & m=5 & Mean \\ \hline
ViT \cite{vaswani2017attention} (m=0)           & 85.8M  & \multicolumn{5}{c}{96.38}                         \\
MLP \cite{almalik2022self} & +625.2M & 94.20 & 96.52 & 95.36 & 95.80 & 95.80 & \textbf{95.54} \\
CNN (Ours) & \textbf{+1.03M} & 93.04 & 96.23 & 96.23 & 94.06 & 94.06 & 94.72\\
ResNet-FT & +13.5M & 95.36 & 93.91 & 93.77 & 90.58 & 90.58 & 92.84\\
ResNet-TL & +2.4M & 93.04 & 92.46 & 93.33 & 91.01 & 92.03 & 92.38\\
ResNet-FT-CNN & +14.2M & 96.38 & 92.75 & 93.19 & 85.80 & 85.80 & 90.78\\
ResNet-TL-CNN & +3.09M & 93.77 & 93.48 & 94.20 & 92.03 & 92.03 & 93.10\\ \hline
\end{tabular}%
}
\end{table}}

\begin{figure}[htpb]
\centering
\begin{minipage}{.5\textwidth}
  \centering
  \includegraphics[width=6cm]{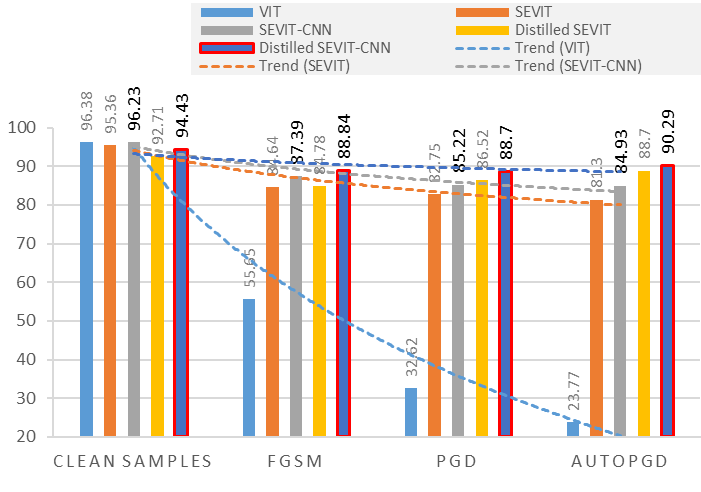}
\end{minipage}%
\hfill
\begin{minipage}{.5\textwidth}
  \centering
  \includegraphics[width=6cm]{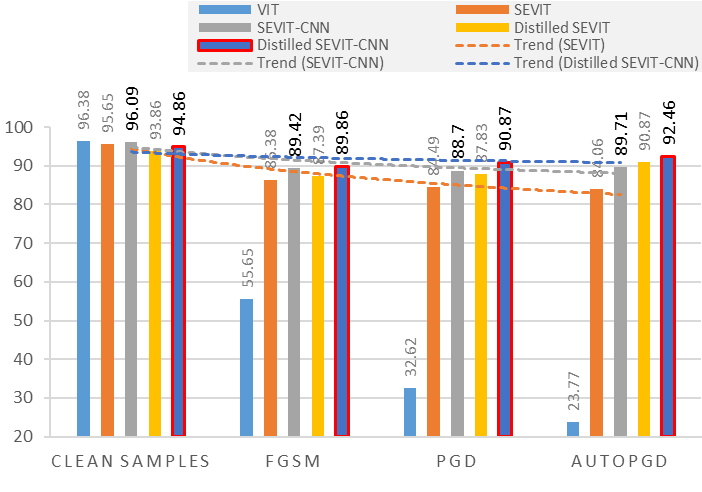}
\end{minipage}
    \caption{Comparative performance of the original ensembling models and their distilled versions in the case of pre-adversarial training (left) vs post-adversarial training (right)}
    \label{fig:dis}
\end{figure}

\begin{figure}[htpb]
\centering
\begin{minipage}{.5\textwidth}
  \centering
  \includegraphics[width=6cm]{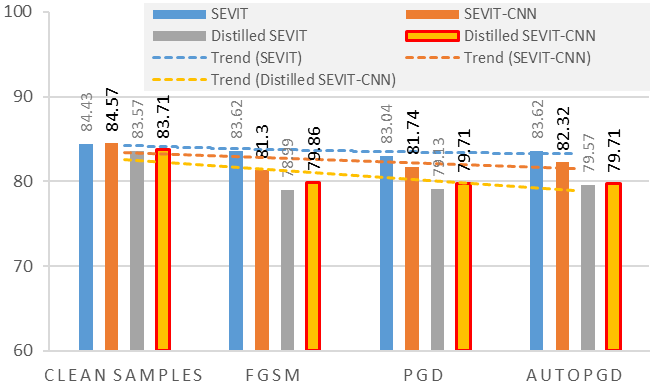}
\end{minipage}%
\begin{minipage}{.5\textwidth}
  \centering
  \includegraphics[width=6cm]{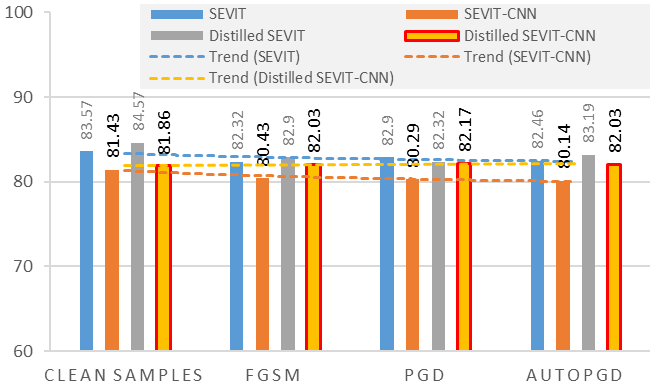}
\end{minipage}
\caption{Comparison of extracted model when model extraction is performed on Dis-
tilled model V/s when extraction is performed on Original (non-distilled) model. The performance of
extracted models are against Clean Samples and Attack Samples. From defender’s view, \textit{Lower} Clean/Adv. Accuracy is better. Pre-adversarial training (left) V/s Post-adversarial training (right)}
\label{fig:ext}
\end{figure}

\end{document}